\documentclass[a4paper, 15pt, journal, twoside]{ieeetran}  % Comment this line out if you need a4paper

\IEEEoverridecommandlockouts                              % This command is only needed if
\usepackage{graphicx}%
\usepackage{subfigure}
\usepackage{multirow}%
\usepackage{amsmath,amssymb}%
\usepackage{amsthm}%
\usepackage{mathrsfs}%
\usepackage{xcolor}%
\usepackage{textcomp}%
\usepackage{booktabs}%
\usepackage{algorithm}%
\usepackage{algorithmicx}%
\usepackage{algpseudocode}%
\usepackage{listings}%
\usepackage[section]{placeins}
\usepackage[justification=centering]{caption}
\usepackage{float}
\usepackage{hyperref}

\title{\LARGE \bf
Disentangled Human Body Representation Based on Unsupervised Semantic-Aware Learning
}

\author{Lu Wang$^{1}$, Xishuai Peng$^{1}$$^{*}$, S. Kevin Zhou$^{2}$% <-this % stops a space

\thanks{$^{1}$Lu Wang and Xishuai Peng are with X-ray Product, Siemens Shanghai Medical Equipment Ltd., No.278, Zhouzhu Rd., Pudong New District, Shanghai, China.
        {\tt\small E-mails: lu.wang.ext@siemens-healthineers.com, xishuai.peng@siemens-healthineers.com}}%
\thanks{$^{2}$S. Kevin Zhou is with University of Science and Technology of China, School of Biomedical Engineering \& Suzhou Institute for Advanced Research Suzhou, Jiangsu, CN.
        {\tt\small E-mails: skevinzhou@ustc.edu.cn}}%
\thanks{$^{*}$Xishuai Peng is the corresponding author
        {\tt\small E-mails: xishuai.peng@siemens-healthineers.com}}
}

\begin{document}

\maketitle
\thispagestyle{empty}
\pagestyle{empty}

%%%%%%%%%%%%%%%%%%%%%%%%%%%%%%%%%%%%%%%%%%%%%%%%%%%%%%%%%%%%%%%%%%%%%%%%%%%%%%%%
\begin{abstract}

In recent years, more and more attention has been paid to the learning of 3D human representation. However, the complexity of lots of hand-defined human body constraints and the absence of supervision data limit that the existing works controllably and accurately represent the human body in views of semantics and representation ability. In this paper, we propose a human body representation with controllable fine-grained semantics and high precison of reconstruction in an unsupervised learning framework. In particularly, we design a whole-aware skeleton-grouped disentangle strategy to learn a correspondence between geometric semantical measurement of body and latent codes, which facilitates the control of shape and posture of human body by modifying latent coding paramerers. With the help of skeleton-grouped whole-aware encoder and unsupervised disentanglement losses, our representation model is learned by an unsupervised manner. Besides, a based-template residual learning scheme is injected into the encoder to ease of learning human body latent parameter in complicated body shape and pose spaces. Because of the geometrically meaningful latent codes, it can be used in a wide range of applications, from human body pose transfer to bilinear latent code interpolation. Further more, a part-aware decoder is utlized to promote the learning of controllable fine-grained semantics. The experimental results on public 3D human datasets show that the method has the ability of precise reconstruction.

\end{abstract}

\begin{IEEEkeywords}
Graph Convolutional Learning, Disentanglement of Body Shape and Posture, Skeleton-grouped Auto-encoder, Representation Learning
\end{IEEEkeywords}

%%%%%%%%%%%%%%%%%%%%%%%%%%%%%%%%%%%%%%%%%%%%%%%%%%%%%%%%%%%%%%%%%%%%%%%%%%%%%%%%
\section{Introduction}
Parameterizing 3D human bodies into low-dimensional representations is crucial for various applications, such as reconstructing and understanding human in computer vision \cite{bogo2016keep, kanazawa2018end, tian2023recovering, zhang2021pymaf}, generating and manipulating character in computer graphics \cite{bouritsas2019neural,tan2018variational,tan2018mesh,yang2014semantic, zeng20173d,zhou2010parametric}. Existing methods \cite{anguelov2005scape,jiang2020disentangled,loper2015smpl,pishchulin2017building, seo2003automatic} usually suffer from limited reconstruction precison because of poor representation capability and coarse semantics. We develop a fine-grained semantic-aware human body models with high precison of reconstruction in this paper.

Since the nonlinear geometric nature of deformations of 3D human bodies caused by poses and shapes, traditional linear representation models \cite{allen2003space,pishchulin2017building,seo2003automatic,yang2014semantic,zeng20173d} cannot accurately reconstruct human body meshes. Therefore, parametric representation models of human meshes like SCAPE \cite{anguelov2005scape}, SMPL \cite{loper2015smpl}, and Adam \cite{joo2018total} have been designed for better capturing 3D mesh deformation with poses and shapes factors. However, their human body shape space is linear to limit the capturing ability of these methods for large deformations of human bodies caused by extremely shape, and hence dissatisfied reconstruction accuracy usually occurs.

With the development of deep learning, the end-to-end encoder-decoder architecture be widely used due to its excellent learning capability \cite{bouritsas2019neural,chen2021learning,gao2021learning,gong2019spiralnet++,ranjan2018generating}. These works extract features from irregular meshes by graph convolution operators to keep reconstruction precision. However, lacking disentangled strategy makes them fail to reach promising reconstruction results when deal with complex human body. Several methods \cite{aumentado2019geometric,chen2021intrinsic,jiang2020disentangled,zhou2020unsupervised} learn disentangle latent code having corresponding coarse semantics, resulting in poor performance on the reconstruction

In this paper, a human body representation model having fine-grained semantics and high reconstruction precision is learned in an unsupervised framework. Two main challenges obstruct the build of the human body model. Firstly, decouping the deformation fator of the human body to controllable fine-grained semantics and high precison of reconstruction is difficult but the key to our work. Secondly, expert knowledge and manual effort for obtain paired supervised human mesh data is very cost, and learning disentangled latent code with keeping reconstruction precision in an unsupervised framework is challenging.

To address above challenges, Disentangled Human Body Representation (DHBR) is proposed, a human body representation model having controllable fine-grained semantics and high precison of reconstruction, which facilitates the control of shape and posture of human body by modifying latent coding paramerers. To learn fine-grained semantics, a whole-aware skeleton-grouped disentangle strategy with anatomical priors of the human skeleton is designed to learn a correspondence between geometric semantical measurement of body and latent codes. Specifically, human body shape variation can be disentangled into whole-body identity-related shape variations (e.g., length, size and style variations) and bone orientation-related variations (e.g.,orientation variations). In contrast to the previous pose and shape disentanglement \cite{anguelov2005scape,jiang2020disentangled,loper2015smpl}, our whole-aware skeleton-grouped disentangle strategy relize a correspondence between pose latent codes and geometric semantical measurement of bone groups and a correspondence between identity latent codes and whole body, which benefits part-level control.

To learn human representation model having fine-grained semantic and high reconstruction precision in unsupervised learning, a skeleton-grouped whole-aware encoder and part-aware decoder architecture are proposed. The part-aware decoder take the fusion of the geometric features of pose information from joints and shape information from whole body to result in precise and efficient learning for human body meshes. Besides, a based-template residual learning scheme is injected into the encoder to ease of unsupervised disentangle learning human body latent parameter and provide geometric regularization in complicated body shape and pose spaces. Experimental results on two public human mesh datasets show the method has the ability of precise reconstruction.

Our main contributions are summarized as follows:
\begin{itemize}
    \item We propose a semantic-aware and disentangled human body representation model, which has controllable fine-grained semantics and high precison of reconstruction. The latent space of our model facilitates the control of shape and posture of human body by modifying latent coding paramerers.
    \item A whole-aware skeleton-grouped disentangle strategy is poposed to learn the priors knowledge of the human body shapes and geometrically meaningful latent coding paramerers in fine-grained semantics.
    \item We propose a skeleton-grouped whole-aware encoder, part-aware decoder architecture and a based-template residual learning scheme to keep the robustness and effectiveness of disentangle learning of shape and pose latent coding paramerers in unsupervised setting.
\end{itemize}

\section{Related work}
\textbf{Statistical Human Parametric Models.}
It is common that human bodies always have strong priors knowledge about its geometric structures. Describing human bodies of statistical parametric models is straightforward. The pioneering works is SCAPE \cite{anguelov2005scape}, modeling variations of different human bodies based on shape-related and pose-related triangle deformations. Classical works is SMPL \cite{loper2015smpl}, representing deformations of human body by vertices offset with more accurately. The statistical parametric models of hands \cite{romero2022embodied}, faces \cite{pavlakos2019expressive}, and animals \cite{zuffi20173d} like SMPL are also explored.

\textbf{Deep Learning of Human models.}
Many convolution-like operators \cite{diederik2014auto,boscaini2015learning,bruna2013spectral,hamilton2017inductive,henaff2015deep,verma2017dynamic} on irregular structures meshes is proposed for deep learning on mesh data. Spectral convolutions \cite{ranjan2018generating} is applied to learn nonlinear embedding of human faces. Spiral convolution \cite{bouritsas2019neural,gong2019spiralnet++} is proposed to learn vertex-based spatial features of meshes. As-consistent-as-possible (ACAP) features-based meshes analysis \cite{tan2018variational,tan2018mesh,jiang2020disentangled,gao2019sparse} are build to represent large deformations of meshes. However, explicit semantics is not considered in these methods and complex geometries are difficult for these operators. Geometric priors knowledge of human bodies is introduced can alleviate the problems in our learning auto-encoder architecture.

\textbf{Disentangled Human Model.}
A deep hierarchical neural network is proposed to learn shape and pose disentangled human model \cite{skafte2019explicit,shu2018deforming,sahasrabudhe2019lifting,rhodin2018unsupervised}, resulting in superior reconstruction precision. However, strong data constraint of paired meshes make the learning cost. Inspired by the unsupervised disentangled strategy \cite{aumentado2019geometric}, various disentangled loss functions \cite{chen2021intrinsic,sun2023learning,zhou2020unsupervised} have been applied. Nevertheless, the reconstruction precision is poor because of not robust disentangled losses. Decoupling pose and shape factor in whole body, disentangled representations can provide only coarse semantics for pose latent codes.

Different from prior works, we propose a whole-aware skeleton-grouped disentangle strategy, which not only has fine-grained semantics but also keeps the robustness and effectiveness of disentangle learning of shape and pose latent coding paramerers in unsupervised setting

\section{Method}

\subsection{Overview of Model}
Given a human mesh dataset with the same topology connectivity structure, our proposed model can capture human body deformation in the view of semantics and precision in an unsupervised setting. In previous unsupervised disentangled models \cite{aumentado2019geometric,chen2021intrinsic,zhou2020unsupervised}, a common solution is that decoupled loss functions are utilized to make models learn shape and pose disentangled representations. However, some methods focus on the whole body when disentanglement, resulting in coarse semantics and not robust enough. Other methods concentrate on the body parts when decoupling, giving rise to lose a large amount of information of whole body and produce some artifacts coupling region between human body parts. In contrast, we adopt a whole-aware skeleton-grouped disentangle strategy (Sec. \ref{dis_str}) to construct the 3D human body model, where fine-grained semantics and high reconstruction-precision are achieved in unsupervised learning.

Based on the disentangle strategy, we introduce a skeleton-grouped whole-aware encoder and part-aware decoder framework (Sec. \ref{encoder}) and the disentangled loss functions (Sec. \ref{loss}) to realize shape and pose disentangled representation.

\subsection{Whole-Aware Skeleton-Grouped Disentanglement} \label{dis_str}
\begin{figure*}[ht]
	\centering
        \includegraphics[width=0.8\textwidth]{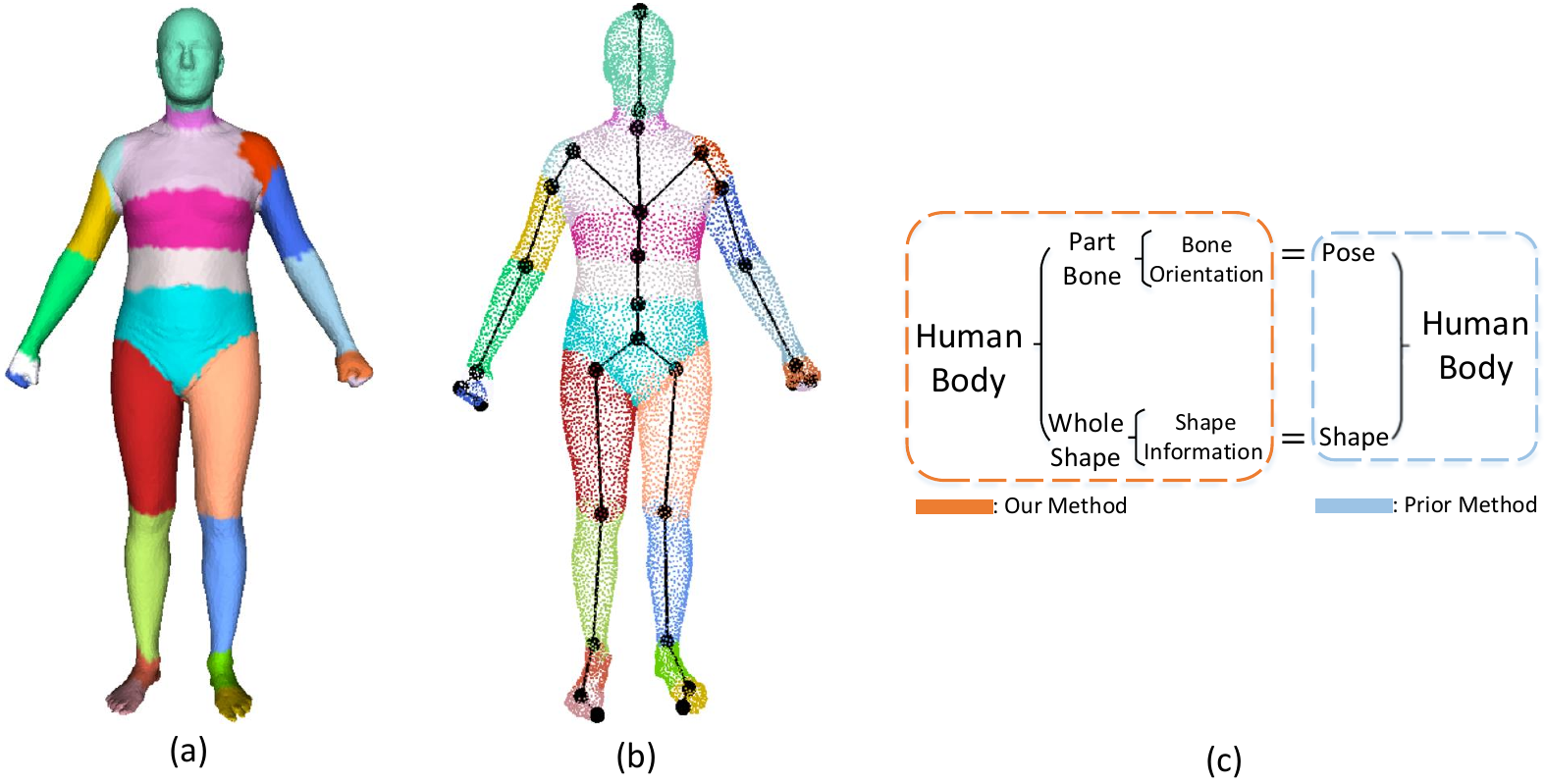}
	\caption{The whole-aware skeleton-grouped disentangle strategy: (a) human body template with anatomical components, (b) human body bones and joints, (c) overview of disentangle strategy.}
	\label{strategy}
\end{figure*}

The human body can be divided into $K = 24$ anatomical components, each including bones and joints defining bone orientation information of human body deformation as shown in Figure \ref{strategy} (a) and (b). Figure \ref{strategy} (c) shows an overview of disentangle strategy of our method. Specifically, human body shape variation can be disentangled into whole-body identity-related shape variations (e.g., length, size and style variations) and bone orientation-related variations (e.g., orientation variations), which are embedded by skeleton-grouped whole-aware encoder as whole-aware identity latent parameter $\boldsymbol{\beta}$ and bone orientation-related pose latent parameter $\boldsymbol{\theta}_k$ for the $k$-th bone group, respectively.

\subsection{Skeleton-Grouped Whole-Aware Encoder} \label{encoder}
\begin{figure*}[ht]
	\centering
        \includegraphics[width=1.0\textwidth]{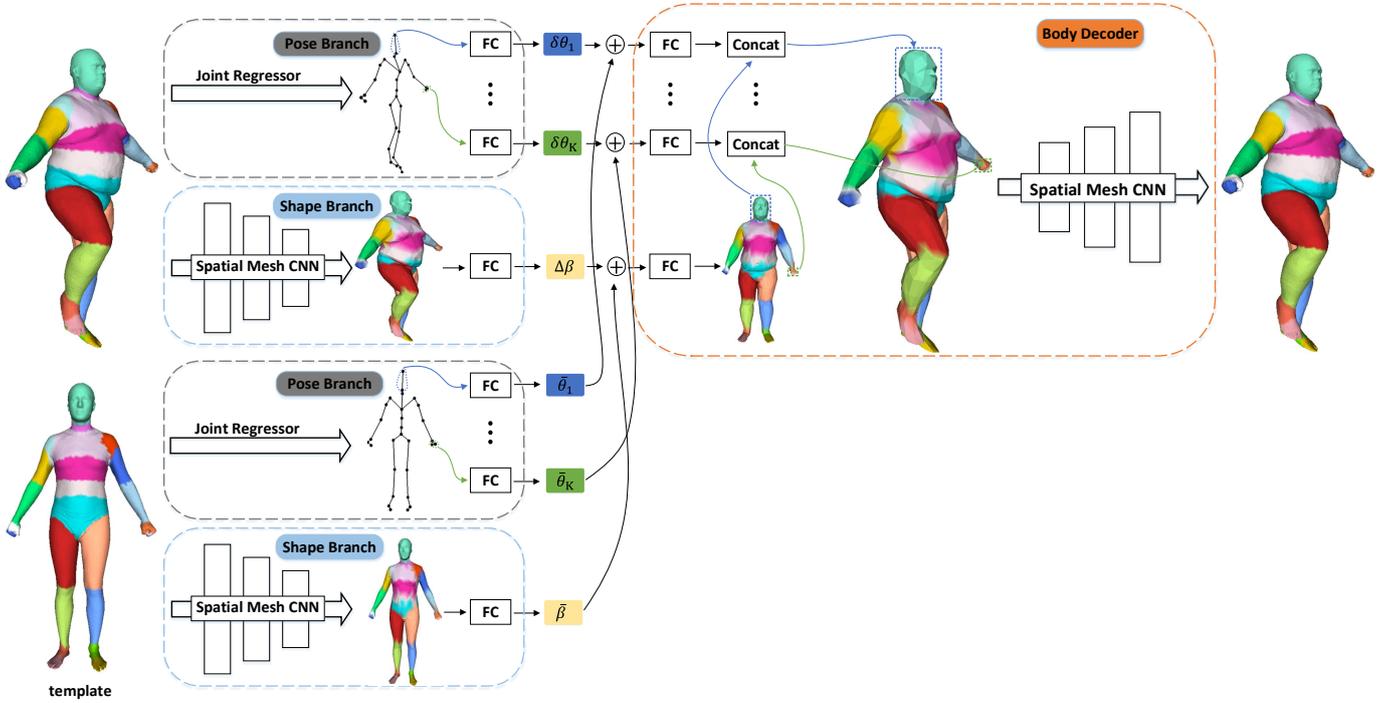}
	\caption{The architecture of our proposed embedding learning network.}
	\label{systemchart}
\end{figure*}

In this section, we describe the pipeline of learning our disentangled human body model in whole-aware skeleton-grouped disentangle strategy. The proposed architecture is shown in Figure \ref{systemchart} in our end-to-end model learning. The encoder consists of two branches for shape $ \boldsymbol{\beta} = E_s(x)$ and for pose $\boldsymbol{\theta} = E_p(x)$ respectively, which are independent and do not share weights. In the encoder, a residual learning scheme is introduce for ease of learning human body latent parameter from complicated body shape and pose spaces. Specially, the human body template $T$ is input into the pose branch and shape branch of encoder $E$ to embed template information as corresponding bone orientation-related pose base latent parameter $ \boldsymbol{\bar{\theta}} =  \left\{ \boldsymbol{\bar{\theta}}_1, \boldsymbol{\bar{\theta}}_2, \ldots, \boldsymbol{\bar{\theta}}_K \right\}$ and whole-aware identity base latent parameter $\boldsymbol{\bar{\beta}}$, respectively. Then given a human body mesh $x$ as input, the pose branch and shape branch of encoder embed the mesh information into the bone orientation-related pose residual latent parameter $ \Delta \boldsymbol{\theta} =  \left\{ \delta \boldsymbol{\theta}_1, \delta \boldsymbol{\theta}_2, \ldots, \delta \boldsymbol{\theta}_K \right\}$ and whole-aware identity residual latent parameter $\Delta \boldsymbol{\beta}$, respectively, where $\delta \boldsymbol{\theta}_k$ is local latent bone orientation-related residual code the $k$-th bone group. As a result, the original pose latent parameter and identity latent parameter of mesh $x$ thus become $\boldsymbol{\theta} = \boldsymbol{\bar{\theta}} + \Delta \boldsymbol{\theta}$ and $\boldsymbol{\beta} = \boldsymbol{\bar{\beta}} + \Delta \boldsymbol{\beta}$, respectively.

In particular, in pose branch, a linear joint regressor $J(\cdot)$ \cite{loper2015smpl} is utilized to generate human body joints $B$ from the input mesh $x$. Then localized bone orientation-related pose residual latent codes $ \left\{ \delta \boldsymbol{\theta}_1, \delta \boldsymbol{\theta}_2, \ldots, \delta \boldsymbol{\theta}_K \right\}$ is inferred from each bone group by different fully connected layers  with $tanh$ as the activation function. Besides, the shape branch consists of hierarchical spiral convolution layers and fully connected layers. Human body mesh $x$ is fed into the spatial spiral convolution layers to generate geometric nonlinear manifold features of plausible shapes space in multiple scales. These nonlinear geometric features of whole-body is taken into subsequently fully connected layers to obtain whole-aware identity residual latent code $\Delta \boldsymbol{\beta}$ encoding global and regional geometric manifold details.

As for decoder, the original pose latent parameter $ \left\{\boldsymbol{\theta}_1, \boldsymbol{\theta}_2, \ldots, \boldsymbol{\theta}_K \right\}$ is fed into different fully connected layers to obtain corresponding bone orientation-related geometric nonlinear manifold features, respectively. Another fully connected layers take the original identity latent parameter $\boldsymbol{\beta}$ as input to predicted whole-aware identity-related geometric nonlinear manifold features. Then these two geometric nonlinear manifold features of vertices are concatenated according to the part labels of vertices. Finally, the spatial spiral convolution layers with similar structure to the shape branch accurately and efficiently reconstructs the original human mesh ${\hat{x}}$ by utilizing the global and local geometric nonlinear manifold features.

\subsection{Loss Functions in Unsupervised Setting} \label{loss}
With the above framework, three loss functions are used to train the end-to-end human model in an unsupervised setting and the overall objective function is defined as:
\begin{equation}
L =  L_{rec} +  L_{c} + L_{s},
\end{equation}
where geometric reconstruction loss $L_{rec}$ could force human model accurately reconstruct human body, disentanglement losses $L_{c}$ and $L_{s}$ could ensure human model decouples bone orientation and shape factors of body. Training process and loss functions will be depicted in detail in the following sections.

\subsubsection{Geometric Reconstruction Loss} \label{geoloss}
To achieve reconstructing the human body mesh as precise as possible, a geometric reconstruction loss is defined as:
\begin{equation}
L_{rec} =  L_{v} + \lambda_{e} L_{e},
\end{equation}
where $\lambda_{e}$ weights the edge regularization. $L_{v}$ is the vertex loss with vertex-wise $L_1$ distance, which ensures the vertices of the reconstructed mesh $D(E_s(x),E_p(x))$ to be as close as possible to the input human body mesh $x$. $L_{v}$ is computed as:
\begin{equation}
L_{v} =  \left \| x - D(E_s(x),E_p(x))\right \|_1.
\end{equation}

Besides, edge regularization $L_{e}$ \cite{groueix20183d,wang2020neural} is introduced to avoid producing over-length edges in $L_{v}$, which could enforce mesh to be tight
to keep the smoothness and reasonableness of reconstructed mesh. $L_{e}$ is calculated as:
\begin{equation}
L_{e} =  \sum_i \sum_{j \in \mathcal{N}(i)} \left \| v_i - v_j\right \|_2^2,
\end{equation}
where $v_i$ is the $i$-th vertex of reconstructed mesh $D(E_s(x),E_p(x))$ and $\mathcal{N}(i)$ is the set of $1$-ring neighbors of vertex $v_i$.

\subsubsection{Unsupervised Disentanglement Losses}
\begin{figure*}[ht]
	\centering
        \includegraphics[width=0.8\textwidth]{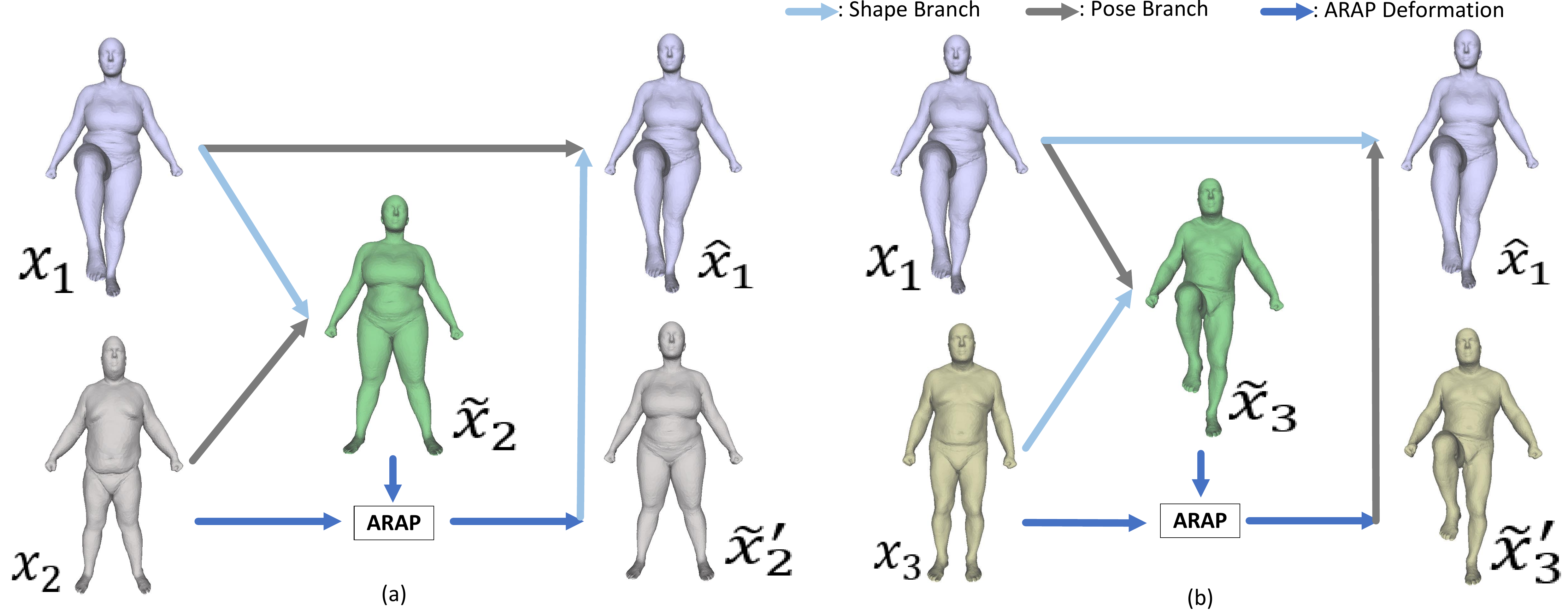}
	\caption{The overview of unsupervised disentanglement losses: (a) the cross-consistency loss, where the shape code of one mesh is utilized to reconstruct itself after a cycle of decoding-encoding process, (b) the self-consistency loss, where the pose code of one mesh is utilized to reconstruct itself after an another cycle of decoding-encoding process.}
	\label{disloss}
\end{figure*}

Accurate reconstruction is already achieved in our human model by  geometric reconstruction loss in Sec. \ref{geoloss}. However, entangled latent space of encoder is still to be addressed in an unsupervised setting. Based on the whole-aware skeleton-grouped decoupling strategy (Sec. \ref{dis_str}), a cross-consistency and a self-consistency loss are utilized to reach this when training model, shown in Figure \ref{disloss}.

Given a mesh triplet $(x_1, x_2, x_3)$, which is randomly sampled from the training set and maybe have different poses and shapes, generating pairs of meshes with two different poses in the exact same shape on the fly when training human model is the key idea behind cross-consistency loss. An intermediate body mesh $\tilde{x}_2 = D(\boldsymbol{\beta}_1, \boldsymbol{\theta}_2)$ is resconstructed from shape latent code from $x_1$ and pose latent code from $x_2$. To avoid degeneracy occurring in trainging, ARAP is utilized to deform $x_2$ to fit the shape information of the decoder reconstructed mesh $\tilde{x}_2$ while keeping the pose information of $x_2$ as much as possible,
\begin{equation}
\tilde{x}_{2}^{'} =  ARAP (x_2, \tilde{x}_2),
\end{equation}
where $\tilde{x}_{2}^{'}$ is the deformed mesh. Specifically, ARAP deformation can be done like \cite{zhou2020unsupervised} by fitting a few anchor points randomly selected from mesh $\tilde{x}_2$. $8\%$ vertices of mesh and $1$ iteration step are used to achieve ARAP deformation. ARAP can work successfully due to the reason that the pose of $\tilde{x}_2$ could converge to the pose of $x_2$ with training. As a result, for pair $(x_2, \tilde{x}_2)$ with only shape information different, the ARAP loss can close to zero. $\tilde{x}_{2}^{'}$ has the pose of $x_2$ with the shape of $x_1$. The cross-consistency loss is defined by swapping shape latent codes of $\tilde{x}_{2}^{'}$ and $x_1$:
\begin{equation}
L_{c} =   \lambda_{c} \left \| D(E_s(\tilde{x}_{2}^{'}),E_p(\mathcal{T}(x_1))) - x_1\right \|_1,
\end{equation}
where $\mathcal{T}$ transforms human meshes when pose fixed (e.g., noise corruption, random scaling), which could make the pose branch has better generalization and robustness. The cross-consistency loss ensure the human model knows the difference shape and pose information. However, part of shape information maybe flow into pose latent codes, resulting in incomplete disentangled latent latent space.

The self-consistency loss is introduced to address this problem. Similar to above, we generate pairs of meshes with two different shapes in the exact same pose on the fly when training. An intermediate body mesh $\tilde{x}_3 = D(\boldsymbol{\beta}_3, \boldsymbol{\theta}_1)$ is resconstructed from shape latent code from $x_3$ and pose latent code from $x_1$. Similar to cross-consistency loss, we use ARAP to deform $x_3$ to match the pose information of the reconstructed mesh $\tilde{x}_3$ while keeping the shape information of $x_3$ as much as possible,
\begin{equation}
\tilde{x}_{3}^{'} =  ARAP (x_3, \tilde{x}_3),
\end{equation}
where $\tilde{x}_{3}^{'}$ is the deformed mesh. For complete disentangled latent space, $x_1$ should be reconstructed from shape latent code from $x_1$ and pose latent code from $\tilde{x}_{3}^{'}$. By minimizing the difference between reconstructed mesh $\hat{x}_1$ and original mesh $x_1$, the self-consistency loss is defined as:
\begin{equation}
L_{s} =   \lambda_{s} \left \| D(E_s(x_1),E_p(\mathcal{T}(\tilde{x}_{3}^{'}))) - x_1\right \|_1,
\end{equation}
where the intuition is that the pose latent code $E_p(\mathcal{T}(\tilde{x}_{3}^{'}))$ must not carry any shape information $E_s(\mathcal{T}(x_3))$.

\subsection{Implementation Details}
Similar to \cite{bouritsas2019neural}, a spiral convolution encoder is utilized in our learning convolution architecture. Specifically, the architecture contains four spiral convolution layers and crossponding downsampling layers, the spiral convolution structure of the decoder is similar to the encoder, only the different part is the upsampling layers replace the the downsampling layers of encoder. The fully-connected layers is placed on the last layer of encoder to map flattened geometric features to latent semantic space. The downsampling and upsampling layers is practiced like \cite{ranjan2018generating} by quadric error metrics. All the training and test experiments are carried out on a PC with an RTX 3060 GPU. We train our human model for $300$ epochs with a learning rate of $5 \times 10^{-3}$, a learning rate decay of $0.9$ after each epoch and the Adam optimizer \cite{kingma2014adam}. About $24$ hours is cost for entire training work. $8$-dimensional latent codes for bone groups and $10$-dimensional latent codes for shape of whole-body are used in our model.

\section{Experiments}
In this section, we evaluate our proposed method on a variety of datasets and tasks. We compare our approach to the state-of-the-art unsupervised human models proposed in \cite{bouritsas2019neural,gong2019spiralnet++,chen2021learning,gao2021learning,sun2023learning}. We also perform an ablation study to evaluate the importance of each part of our model. In addition, we qualitatively show pose transfer and bilinear interpolation results to show the wide applicability of our model.

\subsection{Datasets}
Two publicly available human mesh datasets is used to evaluate our method:

\textbf{SPRING.} The human mesh data in SPRING has the same mesh connectivity as SCAPE \cite{anguelov2005scape}. Yang et al. \cite{yang2014semantic} publishs this large human body dataset in $2014$. There is more than $3,000$ human subject meshes. A non-rigid deformation algorithm is used to register these mesh to a rough A-pose tempalte. Like \cite{sun2023learning}, we randomly split the SPRING dataset into a training set containing $2,743$ human meshes and a test set containing $305$ human meshes.

\textbf{DFAUST.} DFAUST is a large dynamic human body mesh dataset from Bogo et al. \cite{bogo2017dynamic}. The mesh connectivity in DFAUST has the same structure like SMPL \cite{loper2015smpl}. $10$ human subjects are collected in DFAUST. Besides, for each human subject, there is $14$ human body motion 3D sequences to capture different human motion, such as hips, running, jumping , and so on. Like \cite{sun2023learning}, one-twentieth of the original human mesh dataset is sampled to train and test the human model. As a result, we obtain $1,936$ training human meshes and $182$ testing human meshes for learning.

\textbf{Data Preprocessing.} In the learning of our human model, a joint regressor is needed to obtain the joints of human mesh. Besides, body part semantics of each vertex ia also required when decoding the latent codes into recovery huamn mesh. Each dataset has consistent connectivity of mesh structure (it is common for almost all 3D human model learning), we only need align our mesh data with SMPL template once to segment mesh data into different semantical part.

\subsection{Comparison}
In this section, the performance of reconstruction (representation ability) of our human model is evaluated on the SPRING and DFAUST datasets. The average point-wise euclidean distance $E_{avd}$ (in millimeters) calculates the mean euclidean distance from vertices of input human mesh to corresponding vertices of its reconstructed human mesh. This $E_{avd}$ is utilized to evaluate the the performance of reconstruction.

Five kinds of models is compared to validate the accuracy of human body reconstruction: the spiral-based models containing Neural3DMM \cite{bouritsas2019neural} and Spiralplus \cite{gong2019spiralnet++}, the attention-based methods containing Deep3DMM \cite{chen2021learning} and Pai3DMM \cite{gao2021learning}, and the disentanglement models containing SemanticHuman \cite{sun2023learning}. For a fair comparison, the latent space dimension of the official implementation in the compared models is set to the same with our model.

\begin{table*}[h]
	\caption{Quantitative reconstruction performance on SPRING \cite{yang2014semantic} and DFAUST \cite{bogo2017dynamic} datasets. --: not supported for this dataset. Param(M) gives the number of learnable parameters in millions.}
	\centering
    \label{tab:table1}
	\scalebox{1.0}{
    \begin{tabular}{ccccc}
		\toprule
        \multirow{2}*{Methods} & \multicolumn{2}{c}{SPRING} & \multicolumn{2}{c}{DFAUST} \\

        &$E_{avd}$ &Param(M) &$E_{avd}$  &Param(M) \\
		\midrule
		Neural3DMM \cite{bouritsas2019neural}  &6.11  &27.56 &5.49 &30.35 \\
		Spiralplus \cite{gong2019spiralnet++}  &4.99  &13.75 &5.35 &15.15 \\
        Deep3DMM \cite{chen2021learning}       &10.88 &7.84  &9.91 &8.35  \\
        Pai3DMM \cite{gao2021learning}         &4.45  &13.78 &5.76 &15.18 \\
        SemanticHuman \cite{sun2023learning}   &4.33  &1.47  &4.70  &1.59  \\
        DHBR                                   &\textbf{3.62}  &\textbf{1.04}  &\textbf{3.89}  &\textbf{1.23} \\
		\bottomrule
	\end{tabular}}	
\end{table*}

\begin{figure*}[ht]
	\centering
        \includegraphics[width=0.95\textwidth]{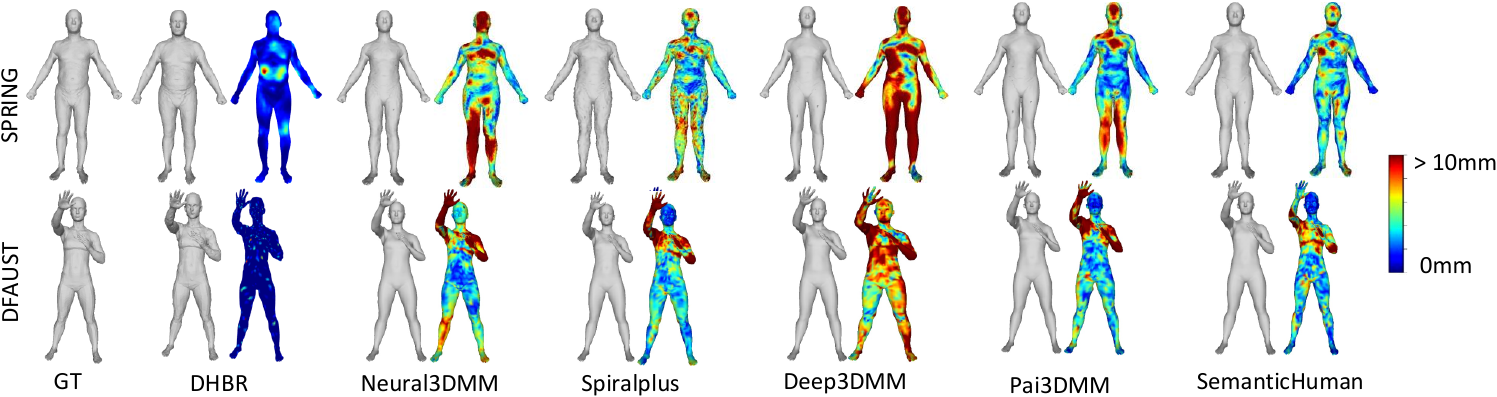}
	\caption{Qualitative reconstruction performance on SPRING \cite{yang2014semantic} and DFAUST \cite{bogo2017dynamic}. The per-vertex Euclidean distance error is encoded as color on the reconstructed meshes. Our human model is trained without any data constraint, which is common in the compared approaches.}
	\label{rec}
\end{figure*}

As shown in Tab. \ref{tab:table1}, our proposed model demonstrates excellent performance to capture human body variations in a small model size. Some reconstruction mesh results and corresponding error maps are shown in Fig. \ref{rec}. We can observed that our human model have better reconstruction accuracy than other approaches, showing the effectiveness of our bone-grouped autoencoder.

\subsection{Ablation Study}
\subsubsection{Effect of Losses}
\begin{table}[h]
	\caption{Quantitative ablation study for reconstruction(in mm). -- : not supported for this task.}
	\centering
    \label{tab:table2}
	\scalebox{1.0}{
    \begin{tabular}{ccccc}
		\toprule
        \multirow{2}*{Methods} & \multicolumn{1}{c}{SPRING} & \multicolumn{1}{c}{DFAUST} &\multirow{2}*{Mean}\\

        &$E_{avd}$  &$E_{avd}$  \\
		\midrule
		Full        &\textbf{3.62} &\textbf{3.89} &\textbf{3.76}  \\
		w/o $L_e$   &4.38 &5.12 &4.75  \\
        w/o $L_c$   &6.43 &7.01  &6.72  \\
        w/o $L_s$   &6.88 &7.69 &7.29  \\
        w/o OLS     &11.45 &12.37  &11.91 \\
		\bottomrule
	\end{tabular}}	
\end{table}

We remove $L_e$, $L_c$, and $L_s$ one by one when training to verify the impact of each loss. As compared in Tab. \ref{tab:table2}, $L_e$ could effectively result in better reconstruction accuracy, and the use of $L_c$ and $L_s$ is required to enable latent space of encoder decoupling. The ablation study give a verification where all loss functions are necessary.

\subsubsection{Effect of Residual Learning Scheme}
We remove residual learning scheme (OLS) in the training process to analyze the impact of it. As compared in Tab. \ref{tab:table2}, we can find that our residual learning scheme results in more better reconstruction meshes by provid strong geometric prior information.

\subsection{Applications}
\subsubsection{Pose transfer}
\begin{figure}[ht]
	\centering
        \includegraphics[width=0.45\textwidth]{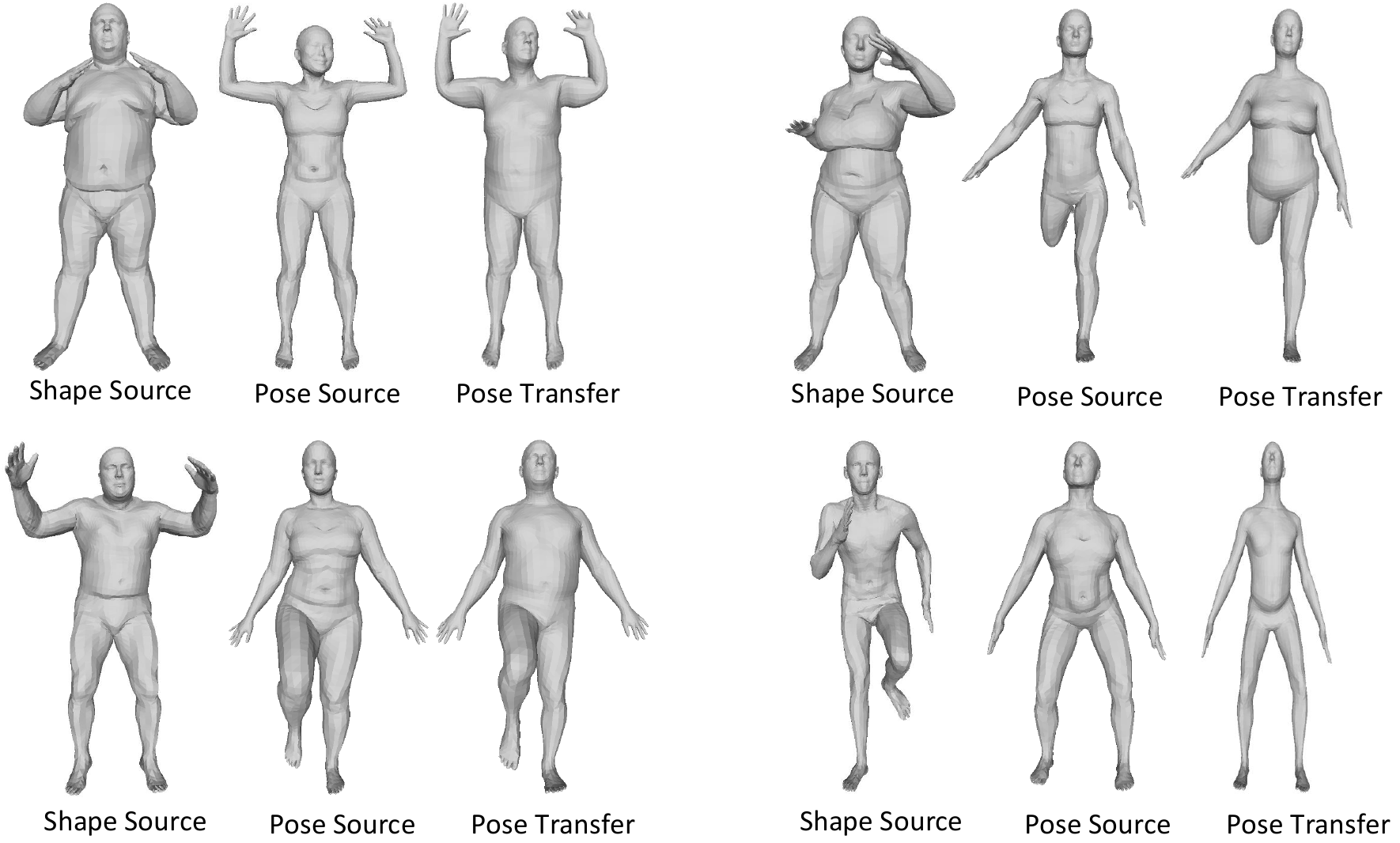}
	\caption{Pose transfer examples from pose source to shape source. In each example, the third new body mesh is decoded from the shape code given by the first mesh and the pose code given by the second mesh.}
	\label{posetransfer}
\end{figure}

For demonstrating the disentanglement of our human representation, the pose transfer is achieved by combining the shape code of a body mesh and the pose code of another body mesh to decode a new body mesh. Fig. \ref{posetransfer} shows examples of pose transfer. It can be found that similar pose and shape information as pose source and shape source are captured by the new body mesh.

\subsubsection{Bilinear Interpolation}
\begin{figure}[ht]
	\centering
        \includegraphics[width=0.45\textwidth]{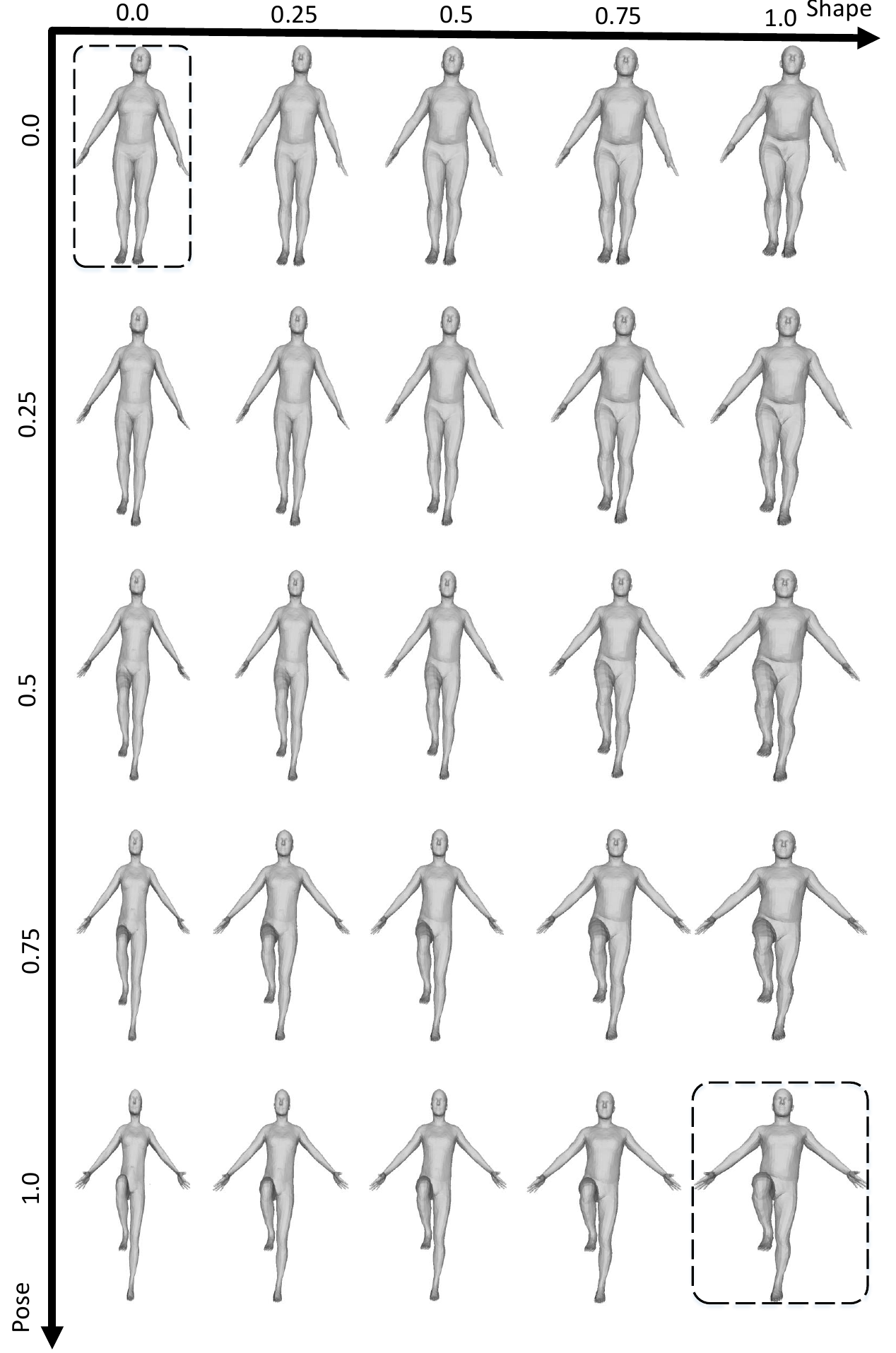}
	\caption{Bilinear interpolation results on the linear latent shape and pose spaces. The meshes in the box are the input reconstructed human bodies.}
	\label{bilinter}
\end{figure}

Our human model decouples the shape and pose factors and encodes them into linear latent spaces, allowing to perform linear interpolation on each kind of latent codes. Given two human meshes having different shapes and poses, their latent shape and pose codes are obtained by encoding these two body mesh, and then the latent shape and pose codes are separately linearly interpolated, as shown in Fig. \ref{bilinter}. It can be observed that the poses in each column are consistent and the shapes in each row are consistent.

\section{LIMITATIONS}
There are several limitations about our work. First, the consistent connectivity is required in our human model training. It is difficult to apply our human model to other mesh data with different mesh connectivity. As a result, we must train differnt human model for different human mesh datasets with different mesh connectivity. It is time consuming and cumbersome.

Second, our human model could not be trained with un-registered point clouds and scan data as input. This limits our human model to know more other type human data except mesh. This problem maybe explored to imporve representation ability of our human model in future work.

\section{Conclusion}
In this paper, a human body representation with controllable fine-grained semantics and high precison of reconstruction is pproposed in an unsupervised learning framework. Specially, a whole-aware skeleton-grouped disentangle strategy is designed to learn a correspondence between geometric semantical measurement of body and latent codes, which facilitates the control of shape and posture of human body by modifying latent coding paramerers. With the help of skeleton-grouped whole-aware encoder and unsupervised disentanglement losses, our representation model is learned by an unsupervised manner. Besides, a based-template residual learning scheme is injected into the encoder to ease of learning human body latent parameter in complicated body shape and pose spaces. Because of the geometrically meaningful latent codes, it can be used in a wide range of applications, from human body pose transfer to bilinear latent code interpolation. Further more, a part-aware decoder is utlized to promote the learning of controllable fine-grained semantics. The experimental results on public 3D human datasets show that the method has the ability of precise reconstruction.

\bibliographystyle{IEEEtran}
\bibliography{IEEEabrv, human_skin_refs}% common bib file

\end{document}